\documentclass[10pt,twocolumn,letterpaper]{article}

\usepackage{cvpr}              

\usepackage{booktabs}
\usepackage{tabularx}
\usepackage{amsmath}
\usepackage{array}
\usepackage{ragged2e}
\usepackage{graphicx}
\usepackage{makecell}
\usepackage{multirow}
\usepackage{float}
\usepackage{caption}
\usepackage{subcaption}

\definecolor{cvprblue}{rgb}{0.21,0.49,0.74}
\usepackage[pagebackref,breaklinks,colorlinks,allcolors=cvprblue]{hyperref}

\def\paperID{5} 
\def\confName{CVPR}
\def\confYear{2026}

\title{TokenDance: Token-to-Token Music-to-Dance Generation \\ with Bidirectional Mamba}

\author{
Ziyue Yang\\
Brown University\\
{\small \texttt{ziyue\_yang@brown.edu}}
\and
Kaixing Yang\\
Renmin University of China\\
{\small \texttt{yangkaixing@ruc.edu.cn}}
\and
Xulong Tang\thanks{Corresponding author.}\\
The University of Texas at Dallas \\
{\small \texttt{xulong.tang@utdallas.edu}}
}

\begin{document}
\maketitle

\begin{abstract}
Music-to-dance generation has broad applications in virtual reality, dance education, and digital character animation. 
However, the limited coverage of existing 3D dance datasets confines current models to a narrow subset of music styles and choreographic patterns, resulting in poor generalization to real-world music.  Consequently, generated dances often become overly simplistic and repetitive, substantially degrading expressiveness and realism. 
To tackle this problem, we present \emph{TokenDance}, a two-stage music-to-dance generation framework that explicitly addresses this limitation through dual-modality tokenization and efficient token-level generation.
In the first stage, we discretize both dance and music using Finite Scalar Quantization, where dance motions are factorized into upper- and lower-body components with kinematic-dynamic constraints, and music is decomposed into semantic and acoustic features with dedicated codebooks to capture choreography-specific structures.
In the second stage, we introduce a Local-Global-Local token-to-token generator built on a Bidirectional Mamba backbone, enabling coherent motion synthesis, strong music-dance alignment, and efficient non-autoregressive inference.
Extensive experiments demonstrate that TokenDance achieves overall state-of-the-art (SOTA) performance in both generation quality and inference speed, highlighting its effectiveness and practical value for real-world music-to-dance applications.
\end{abstract}

\section{Introduction}

With the rapid expansion of internet big data, AIGC tasks have garnered increasing attention from researchers~\cite{zhang2025echomask,zhang2025robust,zhang2025semtalk,zhang2026mitigating}, particularly in the field of AI for Art~\cite{yang2024beatdance}. Dance is an important form of human artistic expression, while music often provides its structural foundation~\cite{mason2012music,li2021ai}. As a result, the Music-to-Dance generation task has emerged, holding broad application prospects~\cite{siyao2022bailando,li2023finedance} in fields such as virtual reality, dance education, and digital character animation.

\begin{figure}[t]
\centering
  \includegraphics[width=0.45\textwidth]{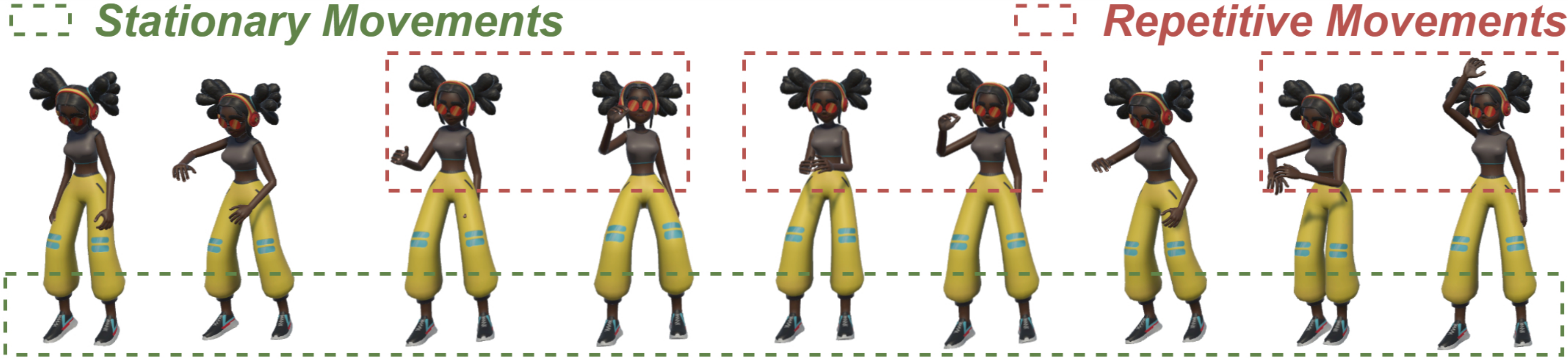}
  \vspace{-0.13in}
  \caption{Generated dance from in-the-wild music.}
  \vspace{-0.25in}
  \label{fig:teaser}
\end{figure}

Existing 3D dance generation methods can be broadly categorized into one-stage and two-stage types. One-stage methods directly regress human motion parameters from audio features~\cite{li2021ai,li2022danceformer,zhuang2022music2dance}. Representative models include Generative Adversarial Network (GAN)-based (CoheDancers~\cite{yang2024cohedancers}), and Diffusion-based models (EDGE~\cite{tseng2023edge}, FineNet~\cite{li2023finedance}, and Lodge~\cite{li2024lodge}). However, as these methods operate in a continuous regression space without an explicit learned motion prior, they are more susceptible to accumulated errors and manifold drift. 
Two-stage methods first construct choreographic units and then learn their probability distributions conditioned on music~\cite{siyao2022bailando,siyao2023bailando++}, simplifying the generation task into a token-based classification problem. By leveraging strong dance priors, these methods enhance motion plausibility, including Bailando~\cite{siyao2022bailando}, Bailando++~\cite{siyao2023bailando++}, MEGADance~\cite{yang2025megadance}. 
However, mainstream 3D dance datasets remain limited in scale, e.g., FineDance~\cite{li2023finedance} with ~8 hours, AIST++~\cite{li2021ai} with ~5 hours, and PopDanceSet~\cite{luo2024popdg} with 3.56 hours. Existing methods either directly regress motion in a continuous space or treat music as a continuous conditioning signal, which weakens explicit modeling of fine-grained rhythmic cues and higher-level musical structure. As a result, they are more prone to repetitive phrases and conservative choreography on structurally complex in-the-wild music, as illustrated in Fig.~\ref{fig:teaser}, leading to reduced expressiveness and realism.

While music appears infinitely diverse from a listener's perspective, from a choreographic standpoint it can be abstracted into a finite set of composable elements~\cite{mason2012music,mitchell2001embodying}. 
Specifically, at the semantic level, dance choreography is typically associated with a limited number of style categories (e.g., Popping, Jazz, Hip-hop), while at the acoustic level, music follows a finite set of rhythmic structures (e.g., 2/4, 3/4, and 4/4 time signatures). 
This observation suggests that, despite surface-level diversity, choreography-relevant musical information lies on a structured and low-dimensional manifold. 
Motivated by this property, we propose to capture such core features through music tokenization-analogous to dance tokenization in~\cite{siyao2022bailando,siyao2023bailando++}-which discovers finite and reusable patterns from complex audio signals and establishes a stable and generalizable prior for dance generation.

Following the above observations, we propose \emph{TokenDance}, a two-stage music-to-dance generation framework designed to explicitly leverage the structured and composable nature of choreography-relevant musical information.
In the \textbf{Dual-Modality Tokenization} stage, both music and dance are discretized using Finite Scalar Quantization (FSQ)~\cite{mentzer2023finite}.
For dance, SMPL~\cite{loper2023smpl} parameters are factorized into upper- and lower-body motions, with kinematic and dynamic constraints applied during reconstruction to ensure physical plausibility.
For music, Librosa-based~\cite{li2021ai} audio representations are decomposed into semantic features and acoustic features, which are quantized using dedicated codebooks to preserve their heterogeneous characteristics.
In the \textbf{Token-to-Token Generation} stage, we introduce a Local-Global-Local generator built upon a Bidirectional Mamba (BiMamba)~\cite{zhu2024vision} backbone.
Specifically, Music Local Scanners independently encode semantic and acoustic music tokens, while a Global Scanner performs joint fusion and refinement with genre-aware modeling.
Finally, two Dance Local Scanners predict upper- and lower-body motion tokens, respectively.
The BiMamba backbone efficiently captures long-range contextual dependencies that are essential for modeling intricate music-dance relationships, leading to improved motion coherence and stronger music-dance alignment.
Moreover, the proposed generator naturally supports non-autoregressive inference, substantially improving generation efficiency.

\begin{figure*}[t]
\centering
  \includegraphics[width=0.85\textwidth]{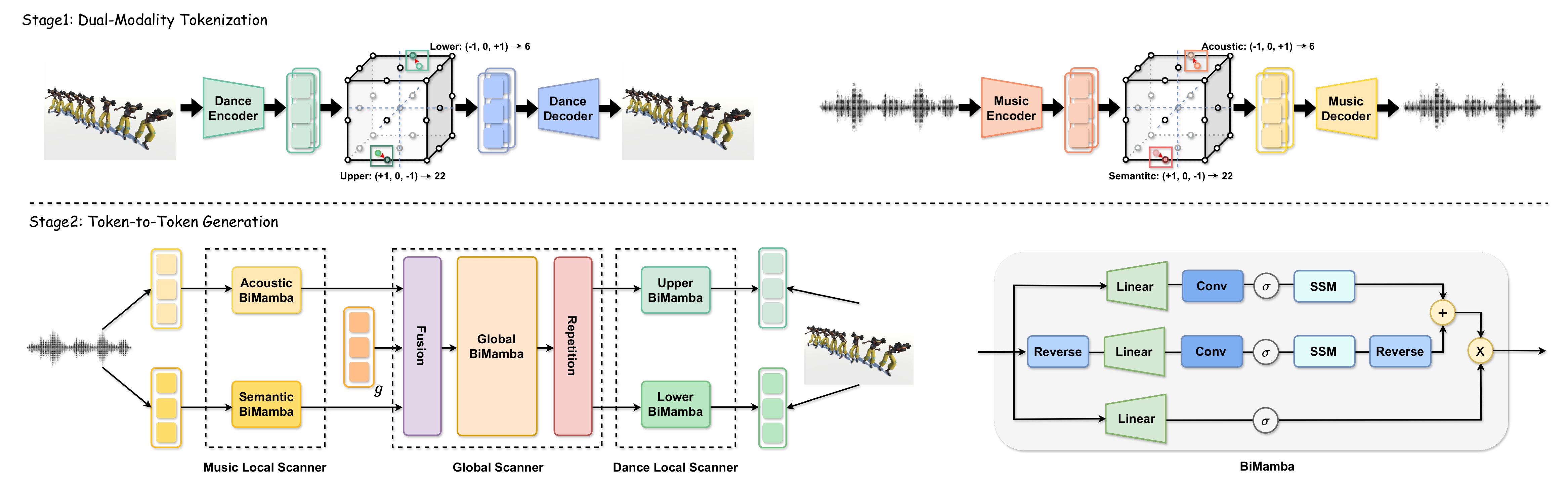}
  \vspace{-0.1in}
  \caption{Overview of TokenDance.}
  \vspace{-0.2in}
  \label{fig:overview}
\end{figure*}

Our contributions to music-to-dance generation are threefold: (1) We propose \emph{TokenDance}, an efficient two-stage music-to-dance generation framework that achieves state-of-the-art (SOTA) performance in both generation quality and inference speed. (2) We introduce \emph{music tokenization} in the dual-modality tokenization stage, which substantially improves model generalization. Extensive experiments further demonstrate its model-agnostic effectiveness. (3) We design a Local-Global-Local token-to-token generator, enabling more coherent motion generation and stronger music-dance alignment. \textit{Additionally, we adopt FSQ for tokenization and BiMamba as the backbone from prior work, while our main contribution lies in dual-modality tokenization and the Local-Global-Local token-to-token generation design.}



\section{Related Work}
\subsection{One-Stage Music-to-Dance Generation}
Music-driven 3D dance generation has attracted increasing attention due to the strong coupling between musical structure and human motion.
Most existing approaches rely on explicit musical representations extracted using audio analysis tools such as Librosa~\cite{li2021ai}, Jukebox~\cite{tseng2023edge}, and MERT~\cite{yang2024codancers}, and aim to predict corresponding human motion representations, including SMPL parameters~\cite{loper2023smpl} or body keypoints~\cite{siyao2022bailando}.

Early studies typically adopt encoder-decoder architectures to directly regress entire motion sequences from music features~\cite{lee2018listen,tang2018dance,le2023music,correia2024music,li2021ai}.
While conceptually simple, such approaches often struggle to capture complex spatial dependencies among human joints.
To address this issue, later works exploit the hierarchical structure of the human body by incorporating Graph Convolutional Networks (GCNs)~\cite{ferreira2021learning,yan2019convolutional}, which explicitly model joint-level interactions and improve biomechanical plausibility.

From a generative modeling perspective, recent advances in AIGC have significantly influenced music-to-dance research.
Generative Adversarial Networks (GANs) have been introduced to enhance motion realism by adversarial supervision, where discriminators guide generators toward more natural and expressive dance motions~\cite{chen2021choreomaster,huang2020dance,yang2024cohedancers}. Recently, Diffusion Models have shown remarkable success in various AIGC tasks, with notable applications extending to the music-to-dance domain~\cite{tseng2023edge,li2023finedance,li2024lodge,li2024lodge++,yang2025flowerdance,yang2025mace}, but the computational cost of the sampling process remains high.

Overall, one-stage methods directly regress motion from music in a continuous space, making them more prone to accumulated errors and manifold drift during inference, especially for long sequences. This motivates two-stage formulations, whose main advantage is the learned discrete latent space or codebook built from real human motion, which provides a stronger motion prior for generation and thus improves motion stability.

\subsection{Two-Stage Music-to-Dance Generation}
Motivated by the inherent periodicity and compositional structure of dance kinematics, two-stage music-to-dance generation methods have been widely explored.
These approaches typically consist of two sequential stages:
(1) a \emph{Dance Quantization} stage, which extracts discrete choreographic units from motion datasets, and
(2) a \emph{Dance Generation} stage, which learns music-conditioned probability distributions over these units.
Since the choreographic units are derived from real human motion data, two-stage methods naturally inherit strong biomechanical priors, leading to improved motion realism and physical plausibility in generated dances.

Early approaches~\cite{ye2020choreonet,aristidou2022rhythm,huang2022genre,chen2021choreomaster} construct choreographic units via uniform temporal segmentation of motion sequences.
While conceptually simple, such strategies incur considerable computational overhead and often fail to capture meaningful motion primitives.
More recent works~\cite{gong2023tm2d} adopt vector-quantized autoencoders (VQ-VAE) to learn discrete motion units in a data-driven manner, significantly reducing time and space complexity while improving reconstruction quality. To further exploit the relative independence between upper- and lower-body motions, ~\cite{siyao2022bailando,siyao2023bailando++} construct separate codebooks for different body parts, effectively expanding the representational capacity of the motion space from $L$ to $L \times L$ and enabling more expressive motion reconstruction.
However, VQ-VAE-based methods often suffer from suboptimal codebook utilization and collapse.
To address this issue, recent studies~\cite{yang2025matchdance,yang2025megadance} introduce Finite Scalar Quantization (FSQ) as an alternative to VQ-VAE, achieving more balanced code usage and improved reconstruction fidelity.
In addition, more sophisticated kinematic and dynamic constraints are incorporated during reconstruction~\cite{yang2025megadance}, enabling accurate modeling of SMPL parameters and surpassing earlier representations based solely on 3D human keypoints~\cite{siyao2022bailando}.

Given discrete choreographic units, the second stage focuses on modeling their music-conditioned temporal distributions.
Early two-stage methods rely on recurrent architectures to capture motion dependencies, such as GRU-based backbones~\cite{chen2021choreomaster} and RNN-based~\cite{huang2020dance} designs.
While effective for short-term modeling, these approaches are limited in capturing long-range musical structure. To address this limitation, more recent methods adopt Transformer-based architectures to enhance temporal reasoning and cross-modal alignment.
~\cite{siyao2022bailando,siyao2023bailando++} introduce cross-modal Transformers that significantly improve music-motion synchronization and global choreographic coherence.
Beyond backbone design, several studies enhance conditioning signals by incorporating explicit genre information.
~\cite{li2023finedance,li2024lodge,zhuang2023gtn} introduce genre cues through lightweight fusion strategies, including cross-attention~\cite{huang2022genre} and feature-level addition~\cite{zhuang2023gtn}.
In parallel, some works explore leveraging large-scale pretrained motion models.
~\cite{gong2023tm2d} adapt text-to-motion pretrained models~\cite{guo2022tm2t} to music-driven settings, achieving improved motion realism but often sacrificing choreographic diversity and creative variability.

However, mainstream 3D dance datasets remain limited in scale, e.g., FineDance~\cite{li2023finedance} with ~8 hours, AIST++~\cite{li2021ai} with ~5 hours, and PopDanceSet~\cite{luo2024popdg} with 3.56 hours. Existing methods either directly regress motion in a continuous space or treat music as a continuous conditioning signal, which weakens explicit modeling of fine-grained rhythmic cues and higher-level musical structure. As a result, they are more prone to repetitive phrases and conservative choreography on structurally complex in-the-wild music, leading to reduced expressiveness and realism.

\section{Methodology}
\subsection{Problem Definition}
Given a genre label $g$ and a music sequence $M=\{m_0, m_1, \ldots, m_T\}$ , our goal is to synthesize the corresponding dance sequence $D=\{d_0, d_1, \ldots, d_T\}$, where $m_t$ and $d_t$ denote the music and dance features at time step $t$. Each $m_t \in \mathbb{R}^{35}$ is extracted with Librosa~\cite{li2024lodge}, and $g$ is encoded as a one-hot vector. Each dance feature $d_t \in \mathbb{R}^{147}$ is represented as $s_t=[\tau;\theta]$, where $\tau$ and $\theta$ denote the root translation and the 6D rotation representation~\cite{zhou2019continuity} of the SMPL parameters~\cite{loper2023smpl}, respectively.

\subsection{Framework Overview}
During training, the \textbf{Dual-Modality Tokenization} stage learns modality-specific codebooks via self-reconstruction with music and dance encoders-decoders, while the \textbf{Token-to-Token Generation} stage trains a generator to map music tokens to dance tokens.

At inference, the first-stage 1D CNN-based \textit{Music Encoder} extracts music tokens, which are then transformed into dance tokens by the second-stage generator, and finally decoded into motion sequence by the first-stage 1D CNN-based \textit{Dance Decoder}.

\subsection{Dual-Modality Tokenization Stage}
\subsubsection{Finite Scalar Quantization.}
Most existing token-based dance generation methods~\cite{siyao2022bailando,siyao2023bailando++} adopt VQ-VAE-style quantization for motion discretization.
However, VQ-VAE often suffers from codebook collapse and uneven code usage, especially when modeling long and complex motion sequences, which limits representational diversity and degrades generation quality~\cite{mentzer2023finite}.

To address these issues, we adopt Finite Scalar Quantization (FSQ)~\cite{mentzer2023finite}, which performs channel-wise discretization without maintaining an explicit codebook.
Specifically, an encoder produces latent features $\mathbf{z}\in\mathbb{R}^d$, which are quantized independently along each channel into discrete indices $\hat{\mathbf{z}} \in \{1,\dots,L\}^d$ using a bounded scalar function (e.g., sigmoid) followed by differentiable rounding:
\begin{equation}
\hat{\mathbf{z}} = f(\mathbf{z}) + \text{sg}\!\left[\text{Round}(f(\mathbf{z})) - f(\mathbf{z})\right],
\end{equation}
where $f(\cdot)$ denotes the bounding function and $\text{sg}$ represents the stop-gradient operator.
The decoder then reconstructs the input signal from $\hat{\mathbf{z}}$.

Unlike VQ-VAE, FSQ does not require auxiliary codebook or commitment losses.
Each channel is constrained to fully utilize all $L$ quantization levels, leading to an effective codebook size of $k=\prod_{i=1}^{d} L_i$.
This design guarantees balanced code utilization by construction, effectively eliminating codebook collapse while maintaining stable gradient propagation during training.

In our implementation, we set $\hat{\mathbf{z}} \in \{8,5,5,5\}$ with $d=4$ channel groups, resulting in a codebook size of $k=1000$.
The models are trained on sequences of 240 frames for 200 epochs using the Adam optimizer with $\beta_1=0.5$ and $\beta_2=0.99$ and with a batch size of 32.

\subsubsection{Dance Tokenization.}
\noindent\textbf{Dance Decomposed Tokenization}. To capture the compositional structure of human motion, we tokenize dance sequences into reusable motion units and construct separate codebooks for the upper and lower body.
This design is motivated by the observation that upper- and lower-body movements often exhibit different temporal patterns and levels of independence in dance choreography.
Decoupling these components allows the model to recombine them more flexibly, thereby enriching motion diversity.
Specifically, a Dance Encoder $\mathbf{E_d}$, consisting of a 3-layer 1D-CNN followed by a 2-layer MLP, encodes the input dance sequence $D=\{D^u, D^l\}$ into context-aware latent features $\mathbf{z}=\{\mathbf{z^u}, \mathbf{z^l}\}$.
These features are quantized using FSQ to obtain discrete representations $\hat{\mathbf{z}}=\{\hat{\mathbf{z^u}}, \hat{\mathbf{z^l}}\}$.
A Dance Decoder $\mathbf{D_d}$, implemented as a 2-layer MLP followed by a 3-layer 1D transposed convolution, reconstructs the motion sequence $\hat{D}=\{\hat{D^u}, \hat{D^l}\}$ from the quantized tokens.

\noindent\textbf{Dynamic-Kinematic Constraint.}
The dance encoder and decoder are trained jointly using a reconstruction objective that enforces consistency in joint positions, velocities, and accelerations, both in joint space and forward kinematics space:
\begin{equation}
\begin{split}
\mathcal{L}_{d} = \ &
\mathcal{L}_{\text{rec}}(\hat{D}, D) 
+ \mathcal{L}_{\text{rec}}(FK(\hat{D}), FK(D)) + \\
& \mathcal{L}_{\text{vel}}(\hat{D}', D') 
+ \mathcal{L}_{\text{vel}}(FK(\hat{D})', FK(D)') + \\
& \mathcal{L}_{\text{acc}}(\hat{D}'', D'') 
+ \mathcal{L}_{\text{acc}}(FK(\hat{D})'', FK(D)''),
\end{split}
\end{equation}
where $'$ and $''$ denote first- and second-order temporal derivatives, respectively, and $FK(\cdot)$ represents the forward kinematics operation~\cite{loper2023smpl}.

\subsubsection{Music Tokenization.}
\noindent\textbf{3D data limitations}. Due to the high cost of acquiring high-quality 3D dance motion, existing datasets remain limited in scale and coverage (e.g., FineDance~\cite{li2023finedance} with ~8 hours, AIST++~\cite{li2021ai} with ~5 hours, PhantomDance~\cite{li2022danceformer} with 9.6 hours, PopDanceSet~\cite{luo2024popdg} with 3.56 hours).
In practice, this often leads to overfitting to frequent audio-motion correlations and poor generalization under diverse or unseen music, manifesting as repetitive and overly conservative dance patterns, as shown in Fig.~\ref{fig:teaser}.

\noindent\textbf{Composable Music Representation.}
At first glance, music appears to exhibit unbounded diversity.
However, from a choreographic perspective, music can be abstracted into a finite set of composable elements~\cite{mason2012music,mitchell2001embodying}. Choreography-relevant musical cues operate at distinct abstraction levels and serve different functional roles.
At the acoustic level, music is governed by a limited set of rhythmic primitives (e.g., 2/4, 3/4, and 4/4 time signatures), which directly constrain motion timing and synchronization.
At the semantic level, choreography is typically associated with a finite number of style categories (e.g., Popping, Jazz, and Hip-hop), which shape motion vocabulary and expressive intent.
Importantly, while surface-level audio realizations may vary continuously, these choreography-relevant cues are drawn from a bounded and repeatedly reused set. Dance-relevant musical information is therefore not uniformly distributed in the raw audio space, but concentrated on a structured and low-dimensional manifold defined by rhythmic regularities and stylistic semantics.

This observation suggests that directly modeling music as a continuous signal is unnecessarily expressive for choreography modeling. Music Tokenization provides a principled way to exploit this structure.
By discretizing continuous audio features into a finite vocabulary, tokenization explicitly constrains the conditioning space to reusable and compositionally meaningful units.
This reformulation transforms music-dance generation from a regression problem into a structured sequence prediction task, which significantly reduces learning complexity under limited data.
As a result, the model is encouraged to reuse learned musical patterns and compose them into novel sequences, yielding improved robustness and generalization to out-of-distribution music.

\noindent\textbf{Music Decomposed Tokenization}.
However, effective music tokenization must respect the multi-level nature of choreographic perception.
Naively collapsing all musical cues into a single discrete space risks entangling heterogeneous factors that operate at different abstraction levels, thereby limiting expressive capacity.

To address this issue, we explicitly decompose music representations into \emph{acoustic} components $\{M^a, \hat{M}^a\}$ and \emph{semantic} components $\{M^s, \hat{M}^s\}$ prior to tokenization. Specifically, the 20-dim MFCC is treated as the semantic component, while the remaining 15 dims are treated as the acoustic component.
This decomposition offers two advantages.
(1) it decouples acoustic-driven and semantic-driven information, preventing mutual interference during discrete modeling.
(2) assigning dedicated FSQ codebooks to each component effectively expands the expressive capacity of the discrete representation from $\mathcal{O}(L)$ to $\mathcal{O}(L^2)$ through compositional combinations of acoustic and semantic tokens.

The Music Encoder $\mathbf{E_m}$ and Decoder $\mathbf{D_m}$ are trained jointly using a reconstruction objective:
\begin{equation}
\mathcal{L}_{m} = \mathcal{L}_{\text{rec}}(\hat{M}^a, M^a) + \mathcal{L}_{\text{rec}}(\hat{M}^s, M^s).
\end{equation}
Through this structured tokenization process, continuous audio signals are mapped into a compact, discrete, and choreography-oriented representation space.
These music tokens provide a stable and generalizable conditioning prior for the subsequent token-to-token generation stage, enabling robust music-conditioned dance synthesis under limited data and diverse musical inputs.

\subsection{Token-to-Token Generation Stage}
\subsubsection{Model Architecture.}
As illustrated in Fig.~\ref{fig:overview}, the Token-to-Token Generation stage adopts a Local-Global-Local architecture built upon a Bidirectional Mamba backbone.
This hierarchical design explicitly separates local temporal modeling from global choreographic reasoning. First, a 2-layer \emph{Music Local Scanner} captures modality-specific temporal dependencies within acoustic and semantic music tokens, focusing on short-range rhythmic and structural cues.
Next, a 4-layer \emph{Global Scanner} integrates information from both modalities and refines the fused representation through a genre-aware gating mechanism.
By conditioning the global representation on genre embeddings, the model enforces stylistic consistency across long motion sequences.
Finally, a 2-layer \emph{Dance Local Scanner} decomposes the global features into upper- and lower-body branches, followed by task-specific classification heads that predict discrete motion tokens. This Local-Global-Local formulation enables TokenDance to jointly model local rhythmic alignment, global choreographic coherence, and body-part-specific motion patterns within a unified framework.

In our implementation, the Mamba block is configured with a model dimension of 512, state size of 16, convolution kernel size of 4, and expansion factor of 2. The model is optimized using Adam with exponential decay rates of 0.9 and 0.99 for the first and second moment estimates, respectively, trained on sequences of 240 frames for 100 epochs with a batch size of 64. Following ~\cite{siyao2022bailando,yang2025megadance}, we train the second-stage generator using a cross-entropy loss over the predicted dance tokens.

\subsubsection{Selective State Space Model.}
In TokenDance, we apply independent Mamba modules to the acoustic-music, semantic music, upper-body motion stream, and lower-body motion stream, respectively, enabling each modality to model its intra-modal temporal dynamics.

\noindent\textbf{High-Quality.} While Transformer-based architectures excel at modeling long-range dependencies, they are inherently position-invariant and rely on positional encodings to capture sequence order~\cite{vaswani2017attention}, which can limit their ability to model fine-grained local continuity.
In contrast, music-to-dance generation critically depends on strong local temporal consistency between successive movements.
Owing to its intrinsic sequential inductive bias, Mamba~\cite{gu2023mamba} has demonstrated superior capability in modeling local dependencies and smooth temporal evolution~\cite{xu2024mambatalk,fu2024mambagesture}.

\noindent\textbf{High-Efficiency.} Moreover, Mamba inherently enables non autoregression sequence generation. Through its parallel scan formulation, Mamba eliminates the need for step-by-step autoregressive decoding, allowing motion sequences to be synthesized in a fully parallel manner. This design substantially improves generation efficiency and scalability, which is particularly critical for 3D dance generation, as inference speed directly impacts interactive scenarios such as real-time motion synthesis, iterative choreography refinement, and user-in-the-loop control.

\noindent\textbf{Mamba.} The Selective State Space Model (Mamba) integrates a selection mechanism with a scan module (S6)~\cite{gu2023mamba} to dynamically emphasize salient input segments during sequence processing.
Unlike classical S4 models~\cite{gu2021efficiently} with fixed state-space parameters $A$, $B$, $C$, and discretization step $\Delta$, Mamba adaptively generates these parameters via fully connected layers, resulting in improved flexibility and generalization.
Formally, for each time step $t$, the input $x_t$, hidden state $h_t$, and output $y_t$ evolve as:
\begin{equation}
\begin{aligned}
h_t &= \bar{A}_t h_{t-1} + \bar{B}_t x_t, \\
y_t &= C_t h_t,
\end{aligned}
\end{equation}
where $\bar{A}_t$, $\bar{B}_t$, and $C_t$ are dynamically predicted.
After discretization with sampling interval $\Delta$, the state transition matrices are given by:
\begin{equation}
\begin{aligned}
\bar{A} &= \exp(\Delta A), \\
\bar{B} &= (\Delta A)^{-1} (\exp(\Delta A) - I) \cdot \Delta B,
\end{aligned}
\end{equation}
with $I$ denoting the identity matrix.
The scan operation efficiently propagates state information across time, allowing the model to capture long sequences with linear complexity.

\subsubsection{Bidirectional Mamba.}
Temporal dependencies in music-to-dance generation are inherently bidirectional.
Musical phrasing often depends on both preceding context and upcoming beats, while choreographic continuity requires anticipating future movements to ensure smooth transitions.
However, the standard Mamba block processes sequences in a unidirectional manner, limiting its ability to leverage future context.

To address this limitation, we introduce \emph{Bidirectional Mamba}, which enhances sequence-wide representations by jointly modeling forward and backward temporal dependencies.
As shown in Fig.~\ref{fig:overview}, the input sequence is processed through a forward Mamba pathway, while a temporally reversed sequence is fed into a backward pathway and subsequently re-inverted.
The outputs from both directions are fused via element-wise addition and further refined through a multiplicative skip connection, which facilitates efficient gradient flow and preserves salient temporal features.
This bidirectional design enables more coherent motion prediction and improves alignment between music structure and generated dance movements.

\begin{table}[t]
  \centering
  \caption{Quantitative analysis on the AIST++ dataset.}
  \vspace{-0.12in}
  \resizebox{0.48\textwidth}{!}{
    \begin{tabular}{lccccc}
      \toprule
      & FID$_k\downarrow$ & FID$_g\downarrow$ & DIV$_k\uparrow$ & DIV$_g\uparrow$ & BAS$\uparrow$ \\
      \midrule
      GT & / & / & 8.19 & 7.45 & 0.2374 \\
      FACT~\cite{li2021ai} 
         & 35.35 & 22.11 & 5.94 & 6.18 & 0.2209 \\
      Bailando~\cite{siyao2022bailando} 
         & 28.16 & \textbf{9.62} & 7.83 & 6.34 & 0.2332 \\
      EDGE~\cite{tseng2023edge} 
         & 42.16 & 22.12 & 3.96 & 4.61 & 0.2334 \\
      Lodge~\cite{li2024lodge} 
         & 37.09 & 18.79 & 5.58 & 4.85 & \textbf{0.2423} \\
      TokenDance 
         & \textbf{21.55} & 11.85 & \textbf{8.05} & \textbf{7.12} & 0.2313 \\
      \bottomrule
    \end{tabular}
  }
  \vspace{-0.1in}
  \label{tab:aist++}
\end{table}

\begin{table}[t]
  \centering
    \caption{Quantitative analysis on the FineDance dataset.}
     \vspace{-0.12in}
  \resizebox{0.48\textwidth}{!}{
    \begin{tabular}{lccccc}
      \toprule
      & FID$_k\downarrow$ & FID$_g\downarrow$ & DIV$_k\uparrow$ & DIV$_g\uparrow$ & BAS$\uparrow$  \\
      \midrule
      GT & / & / & 9.73 & 7.44 & 0.2120  \\
      FACT~\cite{li2021ai} & 113.38 & 97.05 & 3.36 & 6.37 & 0.1831 \\
      Bailando~\cite{siyao2022bailando} & 82.81 & \textbf{28.17} & 7.74 & 6.25 & 0.2029 \\
      EDGE~\cite{tseng2023edge} & 94.34 & 50.38 & \textbf{8.13} & 6.45 & 0.2116 \\
      Lodge~\cite{li2024lodge} & \textbf{45.56} & 34.29 & 6.75 & 5.64 & \textbf{0.2397} \\
      TokenDance & 47.20 & 31.85 & 6.62 & \textbf{6.81} & 0.2385  \\
      \bottomrule
    \end{tabular}
  }
  \label{tab:finedance}
   \vspace{-0.2in}
\end{table}

\section{Experiment}
\subsection{Experimental Setup}
\noindent\textbf{Datasets} 
\textit{(1) AIST++}~\cite{li2021ai} is a widely used benchmark dataset comprising 5.2 hours of 3D street dance motions captured at 60 fps, covering 10 dance genres.
\textit{(2) FineDance}~\cite{li2023finedance} is the largest publicly available dataset for 3D music-to-dance generation, providing 7.7 hours of motion data at 30 fps across 16 distinct dance genres.
\textit{(3) PopDanceSet}~\cite{luo2024popdg} is an in-the-wild dataset collected from Bilibili, comprising 3.56 hours of dance videos across 19 dance styles, and a challenging benchmark for youth-oriented dance generation.

\noindent\textbf{Evaluation Metrics.}
Following prior works~\cite{li2021ai,siyao2022bailando,siyao2023bailando++}, we use FID$_k$ and FID$_g$ to measure motion quality, DIV$_k$ and DIV$_g$ to assess motion diversity, and Beat Alignment Score (BAS) to evaluate rhythmic synchronization.

\subsection{Quantitative Results}
\subsubsection{Generation Quality}
\noindent\textbf{AIST++.}
TokenDance achieves the best overall performance on AIST++ (Tab.~\ref{tab:aist++}), obtaining the lowest FID$_k$ of \textbf{21.55} and the highest motion diversity with DIV$_k$ = \textbf{8.05} and DIV$_g$ = \textbf{7.12}. These results indicate that TokenDance generates high-quality motions while maintaining rich and diverse movement patterns. Although its FID$_g$ (11.85) is slightly higher than that of Bailando (9.62), TokenDance demonstrates a favorable trade-off between motion quality and diversity, validating the effectiveness of its representation and generation strategy.

\noindent\textbf{FineDance.}
On the more challenging FineDance dataset, TokenDance delivers competitive results across most metrics, as shown in Tab.~\ref{tab:finedance}. Specifically, it achieves the lowest FID$_g$ of \textbf{31.85} and the highest DIV$_g$ of \textbf{6.81}, indicating strong global motion consistency and diversity. While Lodge attains the best FID$_k$ (45.56) and BAS (0.2397), TokenDance remains highly competitive with a comparable BAS of 0.2385, suggesting robust rhythmic alignment under complex choreographic settings.

\begin{table}[t]
  \centering
  \caption{Quantitative analysis on the PopDanceSet dataset.}
   \vspace{-0.12in}
  \resizebox{0.48\textwidth}{!}{
    \begin{tabular}{lccccc}
      \toprule
      & FID$_k\downarrow$ & FID$_g\downarrow$ & DIV$_k\uparrow$ & DIV$_g\uparrow$ & BAS$\uparrow$ \\
      \midrule
      GT 
        & / & / & 8.32 & 7.68 & 0.2603 \\
      FACT~\cite{li2021ai} 
        & 37.62 & 26.32 & 5.63 & 6.13 & 0.2162 \\
      Bailando~\cite{siyao2022bailando} 
        & 29.56 & 22.47 & 5.92 & 6.29 & 0.2253 \\
      EDGE~\cite{tseng2023edge} 
        & 34.58 & 23.72 & 6.13 & 6.48 & 0.2334 \\
      POPDG~\cite{luo2024popdg} 
        & 27.13 & 21.41 & \textbf{6.52} & 6.37 & \textbf{0.2403} \\
      TokenDance 
        & \textbf{17.77} & \textbf{15.95} & 6.25 & \textbf{6.94} & 0.2326 \\
      \bottomrule
    \end{tabular}
  }
  \label{tab:popdanceset}
   \vspace{-0.2in}
\end{table}

\noindent\textbf{PopDanceSet.}
As reported in Tab.~\ref{tab:popdanceset}, TokenDance establishes a new state of the art on PopDanceSet in terms of motion quality, achieving the lowest FID$_k$ of \textbf{17.77} and FID$_g$ of \textbf{15.95} among all methods. In addition, TokenDance attains the highest DIV$_g$ of \textbf{6.94}, reflecting superior global motion diversity. Although POPDG slightly outperforms TokenDance in DIV$_k$ (6.52 vs.\ 6.25) and BAS (0.2403 vs.\ 0.2326), TokenDance significantly improves motion realism while maintaining competitive rhythmic alignment across diverse music genres and dance styles.

\subsubsection{Computational Complexity.}
As shown in Table~\ref{tab:latency}, TokenDance achieves the lowest inference latency at both sequence lengths (1.22\,s at 1024 and 2.31\,s at 4096). Compared with prior methods, latency grows much more slowly with sequence length, validating the efficiency and scalability of the non-autoregressive token-to-token design.

\subsection{Qualitative Results}
\begin{figure}[t]
\centering
  \includegraphics[width=0.48\textwidth]{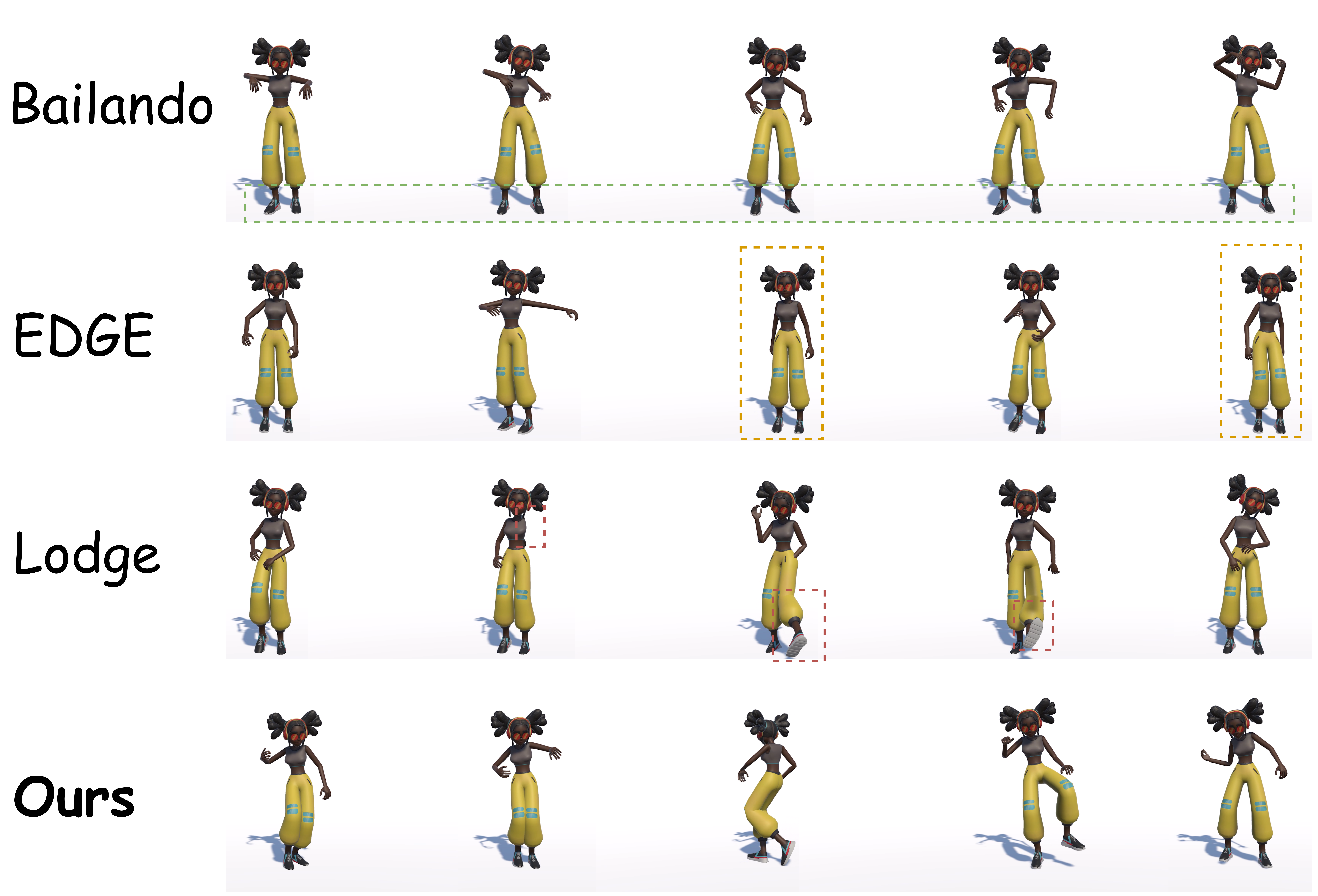}
  \vspace{-0.25in}
  \caption{Qualitative comparison with SOTAs on in-the-wild test samples by models trained on FineDance..}
  \vspace{-0.18in}
  \label{fig:exp}
\end{figure}

\begin{figure}[t]
\centering
  \includegraphics[width=0.48\textwidth]{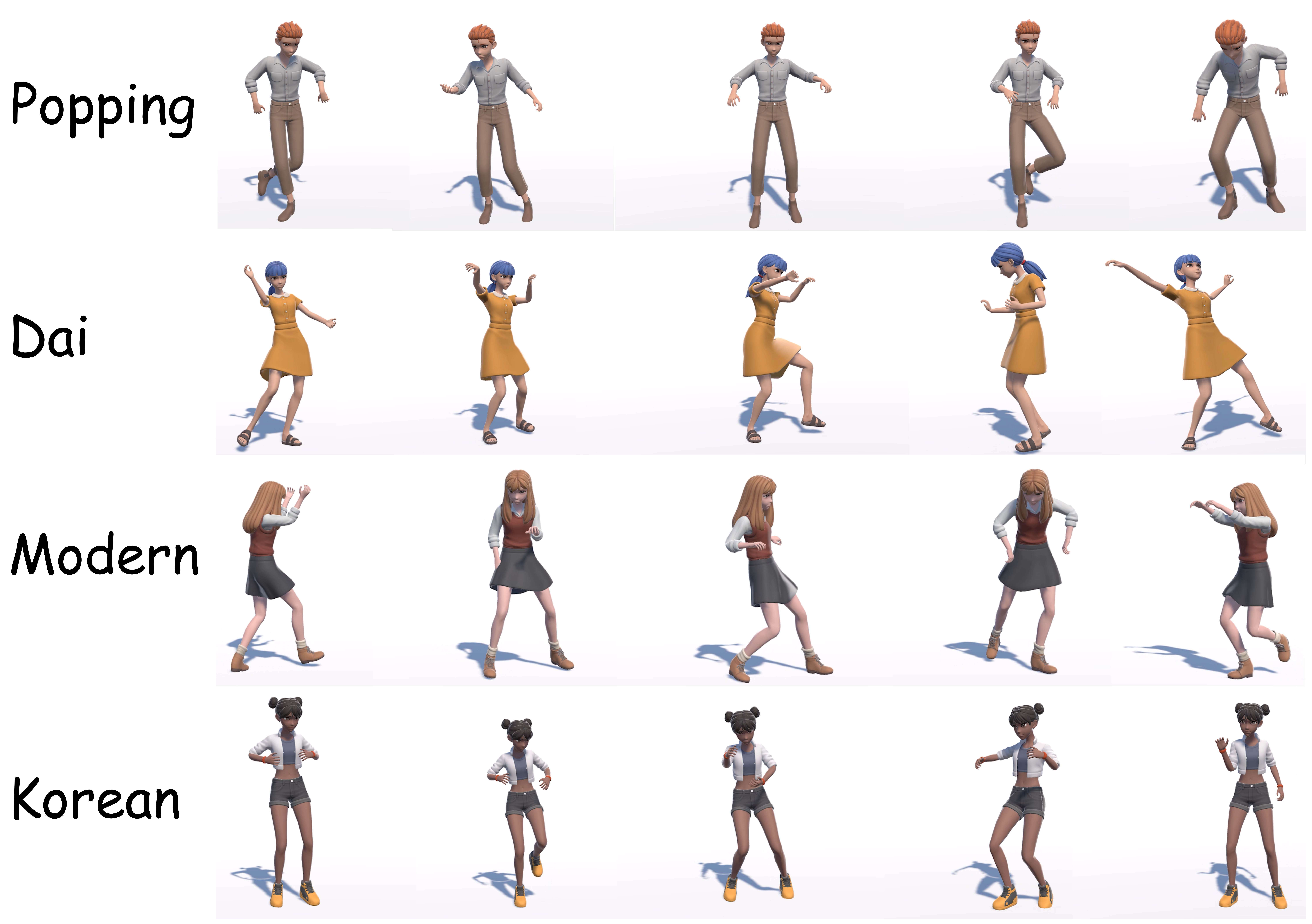}
  \vspace{-0.25in}
  \caption{Qualitative analysis across dance genres on in-the-wild test samples by models trained on FineDance.}
  \vspace{-0.2in}
  \label{fig:genre}
\end{figure}

\textit{Note: All qualitative results in this figure are obtained using models trained on FineDance and evaluated on in-the-wild music.}
\noindent\textbf{(1) Comparison.}
Figure~\ref{fig:exp} shows that TokenDance produces more diverse motions than EDGE~\cite{tseng2023edge}, Lodge~\cite{li2024lodge}, and Bailando~\cite{siyao2022bailando}, with smoother transitions and better perceptual quality. EDGE and Lodge more frequently exhibit repeated short motion loops and occasional unstable foot contacts, while Bailando tends to produce conservative motions with limited spatial coverage. In contrast, TokenDance preserves longer choreographic phrases and more natural upper-lower body coordination.
\noindent\textbf{(2) Cross-Genre.}
As shown in Fig.~\ref{fig:genre}, TokenDance generalizes well across genres (e.g., Dai, modern, popping, and Korean styles), preserving style-specific motion patterns while maintaining music alignment. For example, it captures fluid arm trajectories in Dai dance and sharper isolations in popping without sacrificing temporal smoothness. These observations are consistent with the quantitative improvements in diversity and beat alignment.

\begin{table}[t]
  \centering
  \caption{Comparison on computational latency.}
  \vspace{-0.1in}
  \resizebox{0.45\textwidth}{!}{
    \begin{tabular}{lcc}
      \toprule
      Method & Latency@1024 $\downarrow$ & Latency@4096 $\downarrow$ \\
      \midrule
      FACT~\cite{li2021ai} 
        & 35.88\,s & 142.12\,s \\
      Bailando~\cite{siyao2022bailando} 
        & 5.46\,s & 14.72\,s \\
      EDGE~\cite{tseng2023edge} 
        & 8.59\,s & 27.91\,s \\
      Lodge~\cite{li2024lodge} 
        & 4.57\,s & 11.96\,s \\
      TokenDance 
        & \textbf{1.22\,s} & \textbf{2.31\,s} \\
      \bottomrule
    \end{tabular}
  }
  \label{tab:latency}
  \vspace{-0.2in}
\end{table}

\subsection{User Study}
Following~\cite{legrand2009perceiving,yang2025megadance}, we conduct a double-blind user study with 40 participants on 30 in-the-wild music clips, comparing Bailando~\cite{siyao2022bailando}, EDGE~\cite{tseng2023edge}, Lodge~\cite{li2024lodge}, and TokenDance using 5-point scores on Dance Synchronization (DS), Dance Quality (DQ), and Dance Creativity (DC).
As shown in Table~\ref{tab:user_study}, TokenDance achieves the highest scores across all criteria (DS: $4.12 \pm 0.39$, DQ: $4.09 \pm 0.37$, DC: $3.96 \pm 0.41$), indicating better perceptual quality, synchronization, and creativity. Compared with the strongest baseline Lodge~\cite{li2024lodge}, TokenDance improves DS by 0.41, DQ by 0.31, and DC by 0.27. The remaining gap to GT is also relatively small for DS and DQ, suggesting that the generated motions are close to human choreography in perceived rhythm and realism.

\subsection{Model Agnostic Analysis}
\noindent\textbf{Model Architecture.}
Music Tokenization (MT) is designed to capture choreography-relevant musical structure in a compact and discrete form, independent of the specific model architecture. To verify whether its benefit is model-specific or generalizable, we incorporate MT into three representative baselines with distinct modeling paradigms: FACT~\cite{li2021ai} (one-stage regression), Bailando~\cite{siyao2022bailando} (two-stage token-based generation), and our TokenDance framework. Quantitative results are summarized in Table~\ref{tab:model agnostic}.

Across all three models, MT consistently improves motion quality, diversity, and rhythmic alignment. As shown in Table~\ref{tab:model agnostic}, FACT improves from 113.38/97.05 to 105.12/89.45 on FID$_k$/FID$_g$, Bailando improves from 82.81/28.17 to 68.34/24.50, and TokenDance further improves from 47.20/31.85 to 45.95/30.42.

\noindent\textbf{Music Representation.}
To evaluate the generalization of Music Tokenization (MT) under different music representations, we replace the semantic component (MFCC) in 35-dim Librosa feature of TokenDance with features extracted by MERT~\cite{li2023mert}, Jukebox~\cite{dhariwal2020jukebox}, and Wav2Vec2.0~\cite{baevski2020wav2vec}. We further consider a \emph{w/o. MT} setting, where semantic features are directly concatenated with acoustic features for prediction without tokenization. This design allows us to disentangle the effect of music representation from that of tokenization.

As shown in Table~\ref{tab:ablation_music_representation}, MT consistently improves performance across representations; TokenDance (Full) remains the most balanced setting, while Wav2Vec2.0 is the weakest. These results support that MT is robust to feature choice and that structured tokenization is the key contributor.

\begin{table}[t]
  \centering
  \caption{User study on in-the-wild test samples using models trained on the FineDance dataset.}
  \vspace{-0.1in}
  \resizebox{0.45\textwidth}{!}{
    \begin{tabular}{lccc}
      \toprule
      Method & DS $\uparrow$ & DQ $\uparrow$ & DC $\uparrow$ \\
      \midrule
      GT 
        & $4.52 \pm 0.41$ & $4.45 \pm 0.38$ & $4.37 \pm 0.43$ \\
      FACT~\cite{li2021ai} 
        & $2.11 \pm 0.62$ & $2.03 \pm 0.58$ & $1.98 \pm 0.64$ \\
      Bailando~\cite{siyao2022bailando} 
        & $3.48 \pm 0.51$ & $3.44 \pm 0.49$ & $3.32 \pm 0.53$ \\
      EDGE~\cite{tseng2023edge} 
        & $3.52 \pm 0.47$ & $3.46 \pm 0.50$ & $3.41 \pm 0.48$ \\
      Lodge~\cite{li2024lodge} 
        & $3.71 \pm 0.44$ & $3.78 \pm 0.42$ & $3.69 \pm 0.46$ \\
      TokenDance 
        & $\mathbf{4.12 \pm 0.39}$ & $\mathbf{4.09 \pm 0.37}$ & $\mathbf{3.96 \pm 0.41}$ \\
      \bottomrule
    \end{tabular}
  }
  \label{tab:user_study}
  \vspace{-0.1in}
\end{table}

\begin{table}[t]
  \centering
    \caption{The model-agnostic effect of Music Tokenization (MT) for \textit{model architecture} on the FineDance dataset.}
    \vspace{-0.1in}
  \resizebox{0.49\textwidth}{!}{
    \begin{tabular}{lccccc}
      \toprule
      Method & FID$_k\downarrow$ & FID$_g\downarrow$ & DIV$_k\uparrow$ & DIV$_g\uparrow$ & BAS$\uparrow$ \\
      \midrule
      TokenDance & 47.20 & 31.85 & 6.62 & 6.81 & 0.2385 \\
      TokenDance + MT & \textbf{45.95} & \textbf{30.42} & \textbf{6.75} & \textbf{6.89} & \textbf{0.2412} \\
      \midrule
      Bailando~\cite{siyao2022bailando} & 82.81 & 28.17 & 7.74 & 6.25 & 0.2029 \\
      Bailando~\cite{siyao2022bailando} + MT & \textbf{68.34} & \textbf{24.50} & \textbf{8.01} & \textbf{6.52} & \textbf{0.2147} \\
      \midrule
      FACT~\cite{li2021ai} & 113.38 & 97.05 & 3.36 & 6.37 & 0.1831 \\
      FACT~\cite{li2021ai} + MT & \textbf{105.12} & \textbf{89.45} & \textbf{3.68} & \textbf{6.48} & \textbf{0.1893} \\
      \bottomrule
    \end{tabular}
  }
  \label{tab:model agnostic}
  \vspace{-0.1in}
\end{table}

\begin{table}[t]
  \centering
  \caption{The model-agnostic effect of Music Tokenization (MT) for \textit{music representation} on the FineDance dataset.}
  \vspace{-0.1in}
  \resizebox{0.49\textwidth}{!}{
    \begin{tabular}{lccccc}
      \toprule
      Method & FID$_k\downarrow$ & FID$_g\downarrow$ & DIV$_k\uparrow$ & DIV$_g\uparrow$ & BAS$\uparrow$ \\
      \midrule
      MFCC $\rightarrow$ MERT~\cite{li2023mert} 
        & 42.40 & 45.90 & \textbf{6.62} & 6.78 & 0.232 \\
      MERT~\cite{li2023mert} (w/o. MT) 
        & \textbf{35.10} & 48.60 & 6.42 & 6.50 & 0.228 \\
      \midrule
      MFCC $\rightarrow$ Jukebox~\cite{dhariwal2020jukebox} 
        & 36.80 & 32.46 & 6.15 & 6.53 & 0.202 \\
      Jukebox~\cite{dhariwal2020jukebox} (w/o. MT)  
        & 49.30 & 31.97 & 5.20 & 5.12 & 0.223 \\
      \midrule
      MFCC $\rightarrow$ Wav2Vec2.0~\cite{baevski2020wav2vec} 
        & 60.90 & 32.60 & 6.32 & 5.95 & 0.225 \\
      Wav2Vec2.0~\cite{baevski2020wav2vec} (w/o. MT)  
        & 84.31 & 65.10 & 5.91 & 6.21 & 0.215 \\
      \midrule
      TokenDance (Full) 
        & 47.20 & \textbf{31.85} & \textbf{6.62} & \textbf{6.81} & \textbf{0.239} \\
      \bottomrule
    \end{tabular}
  }
  \label{tab:ablation_music_representation}
  \vspace{-0.1in}
\end{table}

\subsection{Ablation Study}
\noindent\textbf{Music Decomposition.}
We investigate the effect of decomposing music into acoustic and semantic components by removing this design and directly quantizing concatenated music features. As shown in Tables~\ref{tab:music_decomposition_ablation} and~\ref{tab:ablation}, Music Decomposition consistently brings substantial improvements in both music reconstruction and music-to-dance generation tasks. 
In the reconstruction task, introducing decomposition significantly reduces MAE across all levels, with MAE@Full decreasing from 0.544 to 0.345, indicating more accurate and stable modeling of music representations. 
In the downstream generation task, removing Music Decomposition leads to clear performance degradation across all metrics, including higher FID$_k$ (53.12 vs.\ 47.20), higher FID$_g$ (36.45 vs.\ 31.85), and lower diversity and rhythmic alignment. 
These results demonstrate that music decomposition is critical for both faithful reconstruction and downstream dance generation.

\noindent\textbf{Model Backbone.}
We evaluate the impact of the backbone design by replacing the proposed Bidirectional Mamba (BiMamba) with a unidirectional Mamba and a Transformer, as reported in Table~\ref{tab:ablation}. BiMamba achieves the best overall performance. Compared with unidirectional Mamba, BiMamba improves global motion quality and rhythmic alignment, reducing FID$_g$ from 34.20 to 31.85 and increasing BAS from 0.231 to 0.239, while maintaining comparable local motion quality (FID$_k$: 46.95 vs.\ 47.20). In contrast, Transformer performs substantially worse across all metrics, with FID$_k$ rising to 61.37, FID$_g$ to 42.78, and BAS dropping to 0.210.
These results support using BiMamba to jointly model local continuity and global context for better music-dance alignment.

\begin{table}[t]
  \centering
  \caption{Ablation study in music reconstruction task on the FineDance dataset. S, A, and F represent the Semantic, Acoustic, and Full settings, respectively.}
  \vspace{-0.1in}
  \resizebox{0.45\textwidth}{!}{
    \begin{tabular}{lccc}
      \toprule
      Method 
      & MAE@S$\downarrow$ 
      & MAE@A$\downarrow$ 
      & MAE@F$\downarrow$ \\
      \midrule
      w/o. Music Decomp. 
        & 0.642 & 0.358 & 0.544 \\
      TokenDance (Full) 
        & \textbf{0.519} & \textbf{0.269} & \textbf{0.345} \\
      \bottomrule
    \end{tabular}
  }
  \label{tab:music_decomposition_ablation}
  \vspace{-0.1in}
\end{table}

\begin{table}[t]
  \centering
    \caption{Ablation study in music-to-dance generation task.}
    \vspace{-0.1in}
  \resizebox{0.49\textwidth}{!}{
    \begin{tabular}{lccccc}
      \toprule
      Method & FID$_k\downarrow$ & FID$_g\downarrow$ & DIV$_k\uparrow$ & DIV$_g\uparrow$ & BAS$\uparrow$ \\
      \midrule
      w/o. Music Decomp. & 53.12 & 36.45 & 6.10 & 6.22 & 0.225 \\
      BiMamba $\rightarrow$ Mamba & \textbf{46.95} & 34.20 & 6.55 & 6.48 & 0.231 \\
      BiMamba $\rightarrow$ Transformer & 61.37 & 42.78 & 5.75 & 5.89 & 0.210 \\
      TokenDance (Full) & 47.20 & \textbf{31.85} & \textbf{6.62} & \textbf{6.81} & \textbf{0.239} \\
      \bottomrule
    \end{tabular}
  }
  \label{tab:ablation}
  \vspace{-0.1in}
\end{table}

\section{Conclusion}
In this work, we present \emph{TokenDance}, a two-stage music-to-dance generation framework that exploits the structured and composable nature of choreography-relevant musical information. Through Dual-Modality Tokenization, both music and dance are discretized into reusable and semantically meaningful tokens, enabling a more structured formulation of music-to-dance generation under limited 3D dance data. On top of these discrete representations, we introduce a Local-Global-Local token-to-token generator with a Bidirectional Mamba backbone, which jointly captures local rhythmic continuity and global choreographic coherence while enabling efficient non-autoregressive inference. Extensive experiments across multiple datasets show that TokenDance achieves strong overall performance in both generation quality and inference efficiency. Future work will extend the framework with larger and more diverse datasets, as well as text-based conditioning, to enable more flexible and user-specified choreographic control.

{
    \small
    \bibliographystyle{ieeenat_fullname}
    \bibliography{mm-paper}

@inproceedings{siyao2022bailando,
  title={Bailando: 3d dance generation by actor-critic gpt with choreographic memory},
  author={Siyao, Li and Yu, Weijiang and Gu, Tianpei and Lin, Chunze and Wang, Quan and Qian, Chen and Loy, Chen Change and Liu, Ziwei},
  booktitle={Proceedings of the IEEE/CVF Conference on Computer Vision and Pattern Recognition},
  pages={11050--11059},
  year={2022}
}

@article{siyao2023bailando++,
  title={Bailando++: 3d dance gpt with choreographic memory},
  author={Siyao, Li and Yu, Weijiang and Gu, Tianpei and Lin, Chunze and Wang, Quan and Qian, Chen and Loy, Chen Change and Liu, Ziwei},
  journal={IEEE Transactions on Pattern Analysis and Machine Intelligence},
  year={2023},
  publisher={IEEE}
}

@inproceedings{tseng2023edge,
  title={Edge: Editable dance generation from music},
  author={Tseng, Jonathan and Castellon, Rodrigo and Liu, Karen},
  booktitle={Proceedings of the IEEE/CVF Conference on Computer Vision and Pattern Recognition},
  pages={448--458},
  year={2023}
}

@inproceedings{li2024lodge,
  title={Lodge: A coarse to fine diffusion network for long dance generation guided by the characteristic dance primitives},
  author={Li, Ronghui and Zhang, YuXiang and Zhang, Yachao and Zhang, Hongwen and Guo, Jie and Zhang, Yan and Liu, Yebin and Li, Xiu},
  booktitle={Proceedings of the IEEE/CVF Conference on Computer Vision and Pattern Recognition},
  pages={1524--1534},
  year={2024}
}

@inproceedings{yang2024codancers,
  title={CoDancers: Music-Driven Coherent Group Dance Generation with Choreographic Unit},
  author={Yang, Kaixing and Tang, Xulong and Diao, Ran and Liu, Hongyan and He, Jun and Fan, Zhaoxin},
  booktitle={Proceedings of the 2024 International Conference on Multimedia Retrieval},
  pages={675--683},
  year={2024}
}

@inproceedings{le2023music,
  title={Music-driven group choreography},
  author={Le, Nhat and Pham, Thang and Do, Tuong and Tjiputra, Erman and Tran, Quang D and Nguyen, Anh},
  booktitle={Proceedings of the IEEE/CVF Conference on Computer Vision and Pattern Recognition},
  pages={8673--8682},
  year={2023}
}

@inproceedings{li2021ai,
  title={Ai choreographer: Music conditioned 3d dance generation with aist++},
  author={Li, Ruilong and Yang, Shan and Ross, David A and Kanazawa, Angjoo},
  booktitle={Proceedings of the IEEE/CVF International Conference on Computer Vision},
  pages={13401--13412},
  year={2021}
}

@inproceedings{gong2023tm2d,
  title={Tm2d: Bimodality driven 3d dance generation via music-text integration},
  author={Gong, Kehong and Lian, Dongze and Chang, Heng and Guo, Chuan and Jiang, Zihang and Zuo, Xinxin and Mi, Michael Bi and Wang, Xinchao},
  booktitle={Proceedings of the IEEE/CVF International Conference on Computer Vision},
  pages={9942--9952},
  year={2023}
}

@incollection{loper2023smpl,
  title={SMPL: A skinned multi-person linear model},
  author={Loper, Matthew and Mahmood, Naureen and Romero, Javier and Pons-Moll, Gerard and Black, Michael J},
  booktitle={Seminal Graphics Papers: Pushing the Boundaries, Volume 2},
  pages={851--866},
  year={2023}
}

@article{zhuang2022music2dance,
  title={Music2dance: Dancenet for music-driven dance generation},
  author={Zhuang, Wenlin and Wang, Congyi and Chai, Jinxiang and Wang, Yangang and Shao, Ming and Xia, Siyu},
  journal={ACM Transactions on Multimedia Computing, Communications, and Applications (TOMM)},
  volume={18},
  number={2},
  pages={1--21},
  year={2022},
  publisher={ACM New York, NY}
}

@inproceedings{li2023finedance,
  title={Finedance: A fine-grained choreography dataset for 3d full body dance generation},
  author={Li, Ronghui and Zhao, Junfan and Zhang, Yachao and Su, Mingyang and Ren, Zeping and Zhang, Han and Tang, Yansong and Li, Xiu},
  booktitle={Proceedings of the IEEE/CVF International Conference on Computer Vision},
  pages={10234--10243},
  year={2023}
}

@inproceedings{zhou2019continuity,
  title={On the continuity of rotation representations in neural networks},
  author={Zhou, Yi and Barnes, Connelly and Lu, Jingwan and Yang, Jimei and Li, Hao},
  booktitle={Proceedings of the IEEE/CVF conference on computer vision and pattern recognition},
  pages={5745--5753},
  year={2019}
}

@article{li2023mert,
  title={Mert: Acoustic music understanding model with large-scale self-supervised training},
  author={Li, Yizhi and Yuan, Ruibin and Zhang, Ge and Ma, Yinghao and Chen, Xingran and Yin, Hanzhi and Xiao, Chenghao and Lin, Chenghua and Ragni, Anton and Benetos, Emmanouil and others},
  journal={arXiv preprint arXiv:2306.00107},
  year={2023}
}

@inproceedings{yang2024beatdance,
  title={BeatDance: A Beat-Based Model-Agnostic Contrastive Learning Framework for Music-Dance Retrieval},
  author={Yang, Kaixing and Zhou, Xukun and Tang, Xulong and Diao, Ran and Liu, Hongyan and He, Jun and Fan, Zhaoxin},
  booktitle={Proceedings of the 2024 International Conference on Multimedia Retrieval},
  pages={11--19},
  year={2024}
}

@article{dhariwal2020jukebox,
  title={Jukebox: A generative model for music},
  author={Dhariwal, Prafulla and Jun, Heewoo and Payne, Christine and Kim, Jong Wook and Radford, Alec and Sutskever, Ilya},
  journal={arXiv preprint arXiv:2005.00341},
  year={2020}
}

@inproceedings{li2022danceformer,
  title={Danceformer: Music conditioned 3d dance generation with parametric motion transformer},
  author={Li, Buyu and Zhao, Yongchi and Zhelun, Shi and Sheng, Lu},
  booktitle={Proceedings of the AAAI Conference on Artificial Intelligence},
  volume={36},
  number={2},
  pages={1272--1279},
  year={2022}
}

@inproceedings{zhuang2023gtn,
  title={GTN-Bailando: Genre consistent long-term 3d dance generation based on pre-trained genre token network},
  author={Zhuang, Haolin and Lei, Shun and Xiao, Long and Li, Weiqin and Chen, Liyang and Yang, Sicheng and Wu, Zhiyong and Kang, Shiyin and Meng, Helen},
  booktitle={ICASSP 2023-2023 IEEE International Conference on Acoustics, Speech and Signal Processing (ICASSP)},
  pages={1--5},
  year={2023},
  organization={IEEE}
}

@article{vaswani2017attention,
  title={Attention is all you need},
  author={Vaswani, A},
  journal={Advances in Neural Information Processing Systems},
  year={2017}
}

@article{legrand2009perceiving,
  title={Perceiving subjectivity in bodily movement: The case of dancers},
  author={Legrand, Doroth{\'e}e and Ravn, Susanne},
  journal={Phenomenology and the Cognitive Sciences},
  volume={8},
  pages={389--408},
  year={2009},
  publisher={Springer}
}

@article{lee2018listen,
  title={Listen to dance: Music-driven choreography generation using autoregressive encoder-decoder network},
  author={Lee, Juheon and Kim, Seohyun and Lee, Kyogu},
  journal={arXiv preprint arXiv:1811.00818},
  year={2018}
}

@inproceedings{tang2018dance,
  title={Dance with melody: An lstm-autoencoder approach to music-oriented dance synthesis},
  author={Tang, Taoran and Jia, Jia and Mao, Hanyang},
  booktitle={Proceedings of the 26th ACM international conference on Multimedia},
  pages={1598--1606},
  year={2018}
}

@article{chen2021choreomaster,
  title={Choreomaster: choreography-oriented music-driven dance synthesis},
  author={Chen, Kang and Tan, Zhipeng and Lei, Jin and Zhang, Song-Hai and Guo, Yuan-Chen and Zhang, Weidong and Hu, Shi-Min},
  journal={ACM Transactions on Graphics (TOG)},
  volume={40},
  number={4},
  pages={1--13},
  year={2021},
  publisher={ACM New York, NY, USA}
}

@article{huang2020dance,
  title={Dance revolution: Long-term dance generation with music via curriculum learning},
  author={Huang, Ruozi and Hu, Huang and Wu, Wei and Sawada, Kei and Zhang, Mi and Jiang, Daxin},
  journal={arXiv preprint arXiv:2006.06119},
  year={2020}
}

@inproceedings{guo2022tm2t,
  title={Tm2t: Stochastic and tokenized modeling for the reciprocal generation of 3d human motions and texts},
  author={Guo, Chuan and Zuo, Xinxin and Wang, Sen and Cheng, Li},
  booktitle={European Conference on Computer Vision},
  pages={580--597},
  year={2022},
  organization={Springer}
}

@inproceedings{huang2022genre,
  title={Genre-conditioned long-term 3d dance generation driven by music},
  author={Huang, Yuhang and Zhang, Junjie and Liu, Shuyan and Bao, Qian and Zeng, Dan and Chen, Zhineng and Liu, Wu},
  booktitle={ICASSP 2022-2022 IEEE International Conference on Acoustics, Speech and Signal Processing (ICASSP)},
  pages={4858--4862},
  year={2022},
  organization={IEEE}
}

@article{ferreira2021learning,
  title={Learning to dance: A graph convolutional adversarial network to generate realistic dance motions from audio},
  author={Ferreira, Joao P and Coutinho, Thiago M and Gomes, Thiago L and Neto, Jos{\'e} F and Azevedo, Rafael and Martins, Renato and Nascimento, Erickson R},
  journal={Computers \& Graphics},
  volume={94},
  pages={11--21},
  year={2021},
  publisher={Elsevier}
}

@inproceedings{yan2019convolutional,
  title={Convolutional sequence generation for skeleton-based action synthesis},
  author={Yan, Sijie and Li, Zhizhong and Xiong, Yuanjun and Yan, Huahan and Lin, Dahua},
  booktitle={Proceedings of the IEEE/CVF International Conference on Computer Vision},
  pages={4394--4402},
  year={2019}
}

@article{yang2024cohedancers,
  title={CoheDancers: Enhancing Interactive Group Dance Generation through Music-Driven Coherence Decomposition},
  author={Yang, Kaixing and Tang, Xulong and Wu, Haoyu and Xue, Qinliang and Qin, Biao and Liu, Hongyan and Fan, Zhaoxin},
  journal={arXiv preprint arXiv:2412.19123},
  year={2024}
}

@article{li2024lodge++,
  title={Lodge++: High-quality and Long Dance Generation with Vivid Choreography Patterns},
  author={Li, Ronghui and Zhang, Hongwen and Zhang, Yachao and Zhang, Yuxiang and Zhang, Youliang and Guo, Jie and Zhang, Yan and Li, Xiu and Liu, Yebin},
  journal={arXiv preprint arXiv:2410.20389},
  year={2024}
}

@inproceedings{ye2020choreonet,
  title={Choreonet: Towards music to dance synthesis with choreographic action unit},
  author={Ye, Zijie and Wu, Haozhe and Jia, Jia and Bu, Yaohua and Chen, Wei and Meng, Fanbo and Wang, Yanfeng},
  booktitle={Proceedings of the 28th ACM International Conference on Multimedia},
  pages={744--752},
  year={2020}
}

@article{aristidou2022rhythm,
  title={Rhythm is a dancer: Music-driven motion synthesis with global structure},
  author={Aristidou, Andreas and Yiannakidis, Anastasios and Aberman, Kfir and Cohen-Or, Daniel and Shamir, Ariel and Chrysanthou, Yiorgos},
  journal={IEEE Transactions on Visualization and Computer Graphics},
  year={2022},
  publisher={IEEE}
}

@inproceedings{xu2024mambatalk,
  title={Mambatalk: Efficient holistic gesture synthesis with selective state space models},
  author={Xu, Zunnan and Lin, Yukang and Han, Haonan and Yang, Sicheng and Li, Ronghui and Zhang, Yachao and Li, Xiu},
  booktitle={The Thirty-eighth Annual Conference on Neural Information Processing Systems},
  year={2024}
}

@article{correia2024music,
  title={Music to Dance as Language Translation using Sequence Models},
  author={Correia, Andr{\'e} and Alexandre, Lu{\'\i}s A},
  journal={arXiv preprint arXiv:2403.15569},
  year={2024}
}

@article{gu2023mamba,
  title={Mamba: Linear-time sequence modeling with selective state spaces},
  author={Gu, Albert and Dao, Tri},
  journal={arXiv preprint arXiv:2312.00752},
  year={2023}
}

@article{mentzer2023finite,
  title={Finite scalar quantization: Vq-vae made simple},
  author={Mentzer, Fabian and Minnen, David and Agustsson, Eirikur and Tschannen, Michael},
  journal={arXiv preprint arXiv:2309.15505},
  year={2023}
}

@inproceedings{fu2024mambagesture,
  title={MambaGesture: Enhancing Co-Speech Gesture Generation with Mamba and Disentangled Multi-Modality Fusion},
  author={Fu, Chencan and Wang, Yabiao and Zhang, Jiangning and Jiang, Zhengkai and Mao, Xiaofeng and Wu, Jiafu and Cao, Weijian and Wang, Chengjie and Ge, Yanhao and Liu, Yong},
  booktitle={Proceedings of the 32nd ACM International Conference on Multimedia},
  pages={10794--10803},
  year={2024}
}

@article{gu2021efficiently,
  title={Efficiently modeling long sequences with structured state spaces},
  author={Gu, Albert and Goel, Karan and R{\'e}, Christopher},
  journal={arXiv preprint arXiv:2111.00396},
  year={2021}
}

@article{yang2025megadance,
  title={Megadance: Mixture-of-experts architecture for genre-aware 3d dance generation},
  author={Yang, Kaixing and Tang, Xulong and Peng, Ziqiao and Hu, Yuxuan and He, Jun and Liu, Hongyan},
  journal={arXiv preprint arXiv:2505.17543},
  year={2025}
}

@article{yang2025matchdance,
  title={MatchDance: Collaborative Mamba-Transformer Architecture Matching for High-Quality 3D Dance Synthesis},
  author={Yang, Kaixing and Tang, Xulong and Hu, Yuxuan and Yang, Jiahao and Liu, Hongyan and Zhang, Qinnan and He, Jun and Fan, Zhaoxin},
  journal={arXiv preprint arXiv:2505.14222},
  year={2025}
}

@article{yang2025flowerdance,
  title={FlowerDance: MeanFlow for Efficient and Refined 3D Dance Generation},
  author={Yang, Kaixing and Tang, Xulong and Peng, Ziqiao and Zhang, Xiangyue and Wang, Puwei and He, Jun and Liu, Hongyan},
  journal={arXiv preprint arXiv:2511.21029},
  year={2025}
}

@inproceedings{luo2024popdg,
  title={Popdg: Popular 3d dance generation with popdanceset},
  author={Luo, Zhenye and Ren, Min and Hu, Xuecai and Huang, Yongzhen and Yao, Li},
  booktitle={Proceedings of the IEEE/CVF Conference on Computer Vision and Pattern Recognition},
  pages={26984--26993},
  year={2024}
}

@article{mason2012music,
  title={Music, dance and the total art work: choreomusicology in theory and practice},
  author={Mason, Paul H},
  journal={Research in dance education},
  volume={13},
  number={1},
  pages={5--24},
  year={2012},
  publisher={Taylor \& Francis}
}

@article{mitchell2001embodying,
  title={Embodying music: Matching music and dance in memory},
  author={Mitchell, Robert W and Gallaher, Matthew C},
  journal={Music Perception},
  volume={19},
  number={1},
  pages={65--85},
  year={2001},
  publisher={University of California Press}
}

@article{zhu2024vision,
  title={Vision mamba: Efficient visual representation learning with bidirectional state space model},
  author={Zhu, Lianghui and Liao, Bencheng and Zhang, Qian and Wang, Xinlong and Liu, Wenyu and Wang, Xinggang},
  journal={arXiv preprint arXiv:2401.09417},
  year={2024}
}

@article{baevski2020wav2vec,
  title={wav2vec 2.0: A framework for self-supervised learning of speech representations},
  author={Baevski, Alexei and Zhou, Yuhao and Mohamed, Abdelrahman and Auli, Michael},
  journal={Advances in neural information processing systems},
  volume={33},
  pages={12449--12460},
  year={2020}
}

@article{yang2025mace,
  title={MACE-Dance: Motion-Appearance Cascaded Experts for Music-Driven Dance Video Generation},
  author={Yang, Kaixing and Zhu, Jiashu and Tang, Xulong and Peng, Ziqiao and Zhang, Xiangyue and Wang, Puwei and Wu, Jiahong and Chu, Xiangxiang and Liu, Hongyan and He, Jun},
  journal={arXiv preprint arXiv:2512.18181},
  year={2025}
}

@inproceedings{zhang2025echomask,
  title={Echomask: Speech-queried attention-based mask modeling for holistic co-speech motion generation},
  author={Zhang, Xiangyue and Li, Jianfang and Zhang, Jiaxu and Ren, Jianqiang and Bo, Liefeng and Tu, Zhigang},
  booktitle={Proceedings of the 33rd ACM International Conference on Multimedia},
  pages={10827--10836},
  year={2025}
}

@article{zhang2025robust,
  title={Robust 2D skeleton action recognition via decoupling and distilling 3D latent features},
  author={Zhang, Xiangyue and Jia, Yifan and Zhang, Jiaxu and Yang, Yijie and Tu, Zhigang},
  journal={IEEE Transactions on Circuits and Systems for Video Technology},
  year={2025},
  publisher={IEEE}
}

@inproceedings{zhang2025semtalk,
  title={Semtalk: Holistic co-speech motion generation with frame-level semantic emphasis},
  author={Zhang, Xiangyue and Li, Jianfang and Zhang, Jiaxu and Dang, Ziqiang and Ren, Jianqiang and Bo, Liefeng and Tu, Zhigang},
  booktitle={Proceedings of the IEEE/CVF International Conference on Computer Vision},
  pages={13761--13771},
  year={2025}
}

@inproceedings{zhang2026mitigating,
  title={Mitigating Error Accumulation in Co-Speech Motion Generation via Global Rotation Diffusion and Multi-Level Constraints},
  author={Zhang, Xiangyue and Li, Jianfang and Ren, Jianqiang and Zhang, Jiaxu},
  booktitle={Proceedings of the AAAI Conference on Artificial Intelligence},
  volume={40},
  number={15},
  pages={12834--12842},
  year={2026}
}
}

\end{document}